\documentclass[10pt,twocolumn,letterpaper]{article}

\usepackage{wacv}
\usepackage{times}
\usepackage{epsfig}
\usepackage{graphicx}
\usepackage{amsmath}
\usepackage{amssymb}

\usepackage{booktabs}
\usepackage[style=base]{caption}
\usepackage{subcaption}
\usepackage{float}

\DeclareMathOperator*{\argmin}{arg\,min}

\def\wacvPaperID{****} 

\wacvfinalcopy 

\ifwacvfinal
\fi

\ifwacvfinal
\usepackage[breaklinks=true,bookmarks=false]{hyperref}
\else
\usepackage[pagebackref=true,breaklinks=true,colorlinks,bookmarks=false]{hyperref}
\fi

\begin{document}

\title{Ellipse Detection and Localization with Applications to Knots in Sawn Lumber Images}

\author{Shenyi Pan\thanks{Authors contributed equally.}\\
University of British Columbia\\
{\tt\small shenyi.pan@stat.ubc.ca}
\and
Shuxian Fan$^*$\\
University of Washington\\
{\tt\small fansx@uw.edu}
\and
Samuel W.K. Wong\\
University of Waterloo\\
{\tt\small samuel.wong@uwaterloo.ca}
\and 
James V. Zidek\\
University of British Columbia\\
{\tt\small jim@stat.ubc.ca}
\and 
Helge Rhodin\\
University of British Columbia\\
{\tt\small rhodin@cs.ubc.ca}
}

\maketitle

\begin{abstract}
While general object detection has seen tremendous progress, localization of elliptical objects has received little attention in the literature.  Our motivating application is the detection of knots in sawn lumber images, which is an important problem since the number and types of knots are visual characteristics that adversely affect the quality of sawn lumber. We demonstrate how models can be tailored to the elliptical shape and thereby improve on general purpose detectors; more generally, elliptical defects are common in industrial production, such as enclosed air bubbles when casting glass or plastic.  
In this paper, we adapt the Faster R-CNN with its Region Proposal Network (RPN) to model elliptical objects with a Gaussian function, and extend the existing Gaussian Proposal Network (GPN) architecture by adding the region-of-interest pooling and regression branches, as well as using the Wasserstein distance as the loss function to predict the precise locations of elliptical objects. Our proposed method has promising results on the lumber knot dataset: knots are detected with an average intersection over union of 73.05\%, compared to 63.63\% for general purpose detectors.
Specific to the lumber application, we also propose an algorithm to correct any misalignment in the raw lumber images during scanning, and contribute the first open-source lumber knot dataset by labeling the elliptical knots in the preprocessed images.
\end{abstract}

\section{Introduction}\label{sec:intro}
Knots are formed by branches or limbs during the growth of a tree and commonly appear as dark ellipses on the surfaces of sawn lumber.  As they significantly affect both the aesthetic quality and mechanical properties of lumber, knots have a central role in determining the commercial value of lumber.  Such determination is performed by inspecting and measuring the visual characteristics on the surface of the piece.  The cost associated with manual defect detection is prohibitive in industrial fabrication, and thus automated systems are needed.  However, lumber is a highly variable material: sizes and shapes of knots and other defects, as well as color and texture of sawn lumber, have much variability from tree to tree.   The automatic classification of sawn lumber is therefore more challenging than domains where computer vision algorithms are already applied routinely, such as for air bubble detection in glass or plastic casts.


Simple inspection systems that monitor the wood surface with color cameras are already deployed in the lumber industry, exploiting the fact that knots are usually of darker color than the background.  Most commonly, color thresholding is used to mark all knot pixels with a `1'. However, this requires appropriate tuning either by a domain expert or threshold selection methods \cite{otsu1979threshold}. As a result, the performance of these detection methods is highly sensitive to the choice of threshold and associated geometric features selected for a particular wood type and camera setup. Moreover, ellipse detection in imperfectly binarized images remains a difficult and error-prone problem \cite{libuda2007ellipse,wang2014fast,chia2007ellipse,muammar1991tristage,tsuji1978detection,aguado1995ellipse}. We provide a comparison of our proposed deep-learning-based ellipse detection method to one of the geometric detection methods in \cite{jia2017fast} and to other learning-based variants on the lumber knot images in the experiment section.

\begin{figure}
	\centering
	\includegraphics[width= 0.47\textwidth]{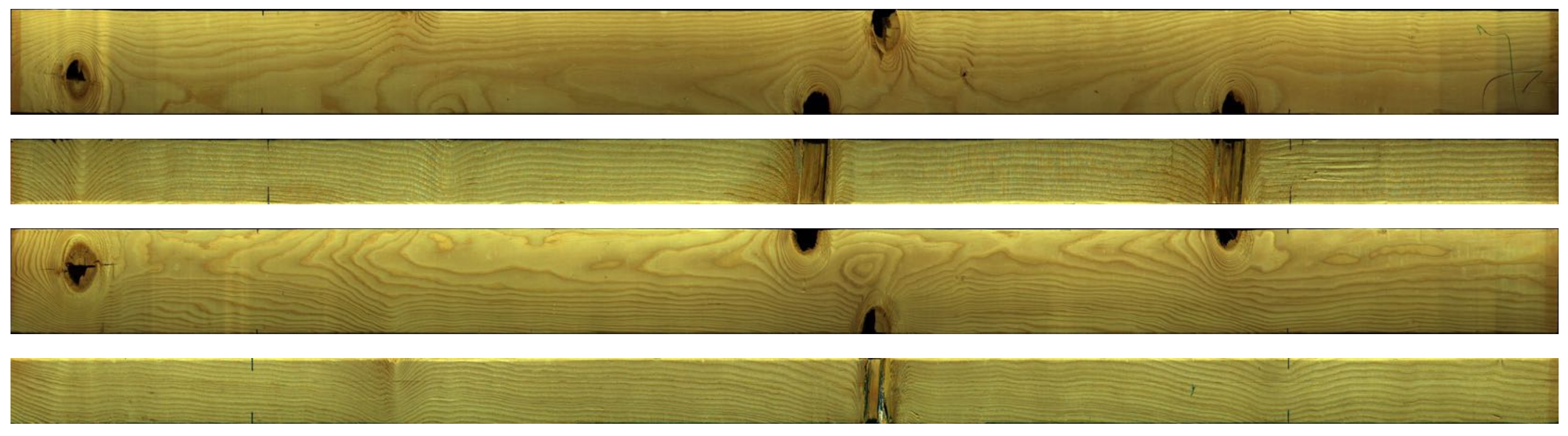}
	\caption{Sample lumber with the four surfaces scanned. The wide surfaces are shown on the first and the third rows and the narrow surfaces are shown on the second and last rows. Knots can be seen from their darker color and the noticeable grain distortion around them.}
	\label{fig: knot_example}
\end{figure}

In the literature, knots are typically modeled as elliptical cones when a piece of lumber is treated as a 3-dimensional object \cite{guindos2013three}. Hence, we model knot faces on a 2-dimensional surface are as ellipses (conic sections). The proposed knot detection method takes color images of lumber captured by high-definition cameras as the input and returns a 5-dimensional parameter vector, $(cx, cy, rx, ry, \theta)$, representing the position of the center of the knot face on the $x$-axis and $y$-axis, the length of the semi-diameters along the $x$- and $y$-axes, along with the rotation angle of the ellipse $\theta$, respectively.
The available data for this paper consist of images of 113 large sawn Douglas-fir lumber specimens with 4894 knots in total. The data are collected in a collaborative research project between the Department of Statistics at the University of British Columbia and FPInnovations. Figure~\ref{fig: knot_example} shows an example of knots on a piece of sawn lumber. 
Four images are taken for each board: two for the wide surfaces and another two for the narrow surfaces. 
For training and testing purposes, we manually labeled each knot.




In this paper, we address the knot detection and localization problem by adapting the Faster R-CNN framework \cite{ren2015faster} for rectangular object detection. Our main contributions are as follows:
\begin{itemize}
	\item We adapt the Faster R-CNN to identify elliptical knots on the sawn lumber surfaces. Specifically, we replace the Region Proposal Network in the Faster R-CNN framework with the Gaussian Proposal Network to adapt to the ellipse detection problem, and propose an alternative loss function for ellipse regression.
	\item We propose an effective algorithm to fix the misalignment of raw images of lumber specimens.
	\item We prepare a labeled dataset of elliptical lumber knots, which can be used by future researchers to train and evaluate methods for knot detection and localization. 
\end{itemize}

Knot detection is vital for assessing the strength and value of sawn lumber in a non-destructive manner. In particular, knot detection and localization based on lumber images is an integral step that generates input for existing knot matching algorithms that reconstruct the 3-dimensional structure of knots, e.g., \cite{jun2019sequential}.  Thus, this paper provides an important step in the pipeline towards advancing the state-of-the-art in automatic strength prediction of lumber and quality assessment of other materials containing elliptical defects. 

This paper is organized as follows. Related work regarding lumber knot localization and elliptical object detection is reviewed in Section~\ref{sec:relatedwork}. The data preparation procedures and the lumber knot dataset are discussed in Section~\ref{sec:dataprep}. Section~\ref{sec:approach} introduces our proposed method to solve the knot detection and localization problem.  Experiment results and model performance evaluations are presented in Section~\ref{sec:exp}. Section~\ref{sec:conc} concludes this paper.

\section{Related Work}
\label{sec:relatedwork}

In this section, we discuss existing lumber defect detection methods, including active sensor-based solutions and vision-based solutions using geometry, machine learning (ML), and combinations thereof.

\paragraph{Active, using the laser tracheid effect.}
An existing method for knot location and size measurements in veneer (a thin layer of wood) surfaces using the laser tracheid effect is introduced in \cite{tormanen2009detection}. This method is based on the scattering patterns formed by laser spots. The laser light that penetrates the lumber surface is scattered mainly in the grain direction; this is often referred to as the ``tracheid effect''. The major features used to detect knots are the deviation of wood grain obtained by analyzing the amount of scattering. Areas with knots generally indicate the existence of large wood grain deviations and can be detected by thresholding. Methods based on the ``tracheid effect'' focus on finding improved approaches to computing the grain deviation. For example, Daval \textit{et al.}\cite{daval2015automatic} use the thermal conduction properties of lumber captured by a thermal camera to extract information such as the slope of grain and the presence of knots. The performance of the tracheid effect-based methods depends on the analysis of multiple individual laser spots projected on the surface. A practical limitation is in the resolution at which surface characteristics are captured. Empirical results show that tracheid effect-based methods are incapable of identifying knots of smaller sizes since they are usually located between adjacent laser spots. Therefore, these small knots are difficult to detect as they do not cause sufficiently large changes in the scattering patterns.

\paragraph{Passive geometric localization.}
Computer vision-based ellipse detection methods often use a multi-stage filtering process, by finding geometric features, such as lines, curves, arcs, and extended arc patterns as intermediate representations~\cite{libuda2007ellipse,teutsch2006real,kim2002fast,mai2008hierarchical,chia2010split,prasad2010ellipse,chia2011object,fornaciari2014fast,prasad2014deb,jia2017fast,dong2018accurate}. Improvements on the contour extraction and arc combination algorithms have been recognized in many real-world applications~\cite{wang2014fast,lu2019arc,jin2019ellipse}. Previous work applied to knot detection used image processing, morphological processing, and feature extraction procedures to detect sizes and locations of knots on lumber surfaces \cite{todoroki2010automated,yang2017}. Global thresholding is used to segment images through morphological operations to isolate regions that are likely to contain knots. Adaptive thresholding is then applied to suspected areas to improve the accuracy of knot segmentation results. However, the lack of robustness to noise and blurry shape edges still persists as major limitations to knot detection applications. The shape detection accuracy often deteriorates substantially when noise and blurriness of knots edges increase.

\paragraph{ML-based detection and localization.}
To identify and detect the locations of knots based on images, an alternative to modeling the knots as ellipses is to use object detection methods to find bounding boxes around knots. Bounding box regression and object detection have been a long-standing problem in the field of computer vision. The aim of bounding box regression is to refine or predict the minimum localization boxes within which objects of interest lie. With the developments in deep learning in recent years, many effective methods have been proposed to detect objects from an image. Methods such as R-CNN \cite{girshick2014rich}, Fast R-CNN \cite{girshick2015fast}, and Faster R-CNN \cite{ren2015faster} apply different methods to select possible regions containing objects. In particular, Faster R-CNN implements a Region Proposal Network (RPN) to propose plausible regions that are likely to contain objects. YOLO \cite{redmon2016you} segments images into smaller pieces and detects objects within and across the smaller pieces. SSD \cite{liu2016ssd} is another single shot detection method similar to YOLO, but it uses feature maps from different layers to detect objects of different sizes. However, these object-detection methods are proposed to use bounding boxes to detect rectangular objects and have limitations when operating on elliptical objects with inherent symmetries.

\paragraph{Elliptical object localization.}
More recently, papers have been proposed to detect elliptical objects using machine learning-based methods. For example, 
Dong \textit{et al.} \cite{dong2020ellipse} propose Ellipse R-CNN to detect clustered and occluded ellipses and suggest an objective function for performing ellipse regression. However, their method has a complicated pipeline which includes many image cropping and scaling operations tailored for detecting occluded ellipses, which is not the case for lumber knots. Moreover, Wang \textit{et al.} \cite{wang2019ellipse} propose an Ellipse Proposal Network to detect optic boundaries in fundus images. However, the technical details regarding the objective functions and network architecture were not elaborated in the paper. Another application to pupil localization based on region proposal network and Mast R-CNN is proposed with a new computational schedule of anchor ellipse regression in \cite{lin2019pupil}. Li \cite{li2019detecting} proposes a Gaussian Proposal Network (GPN) to replace the usual region proposal network in object detection frameworks such as Faster R-CNN. The elliptical proposals are parameterized as Gaussian distributions and the GPN is trained to minimize the Kullback-Leibler (KL) divergence between the ground truths and the proposed ellipses. However, the proposed GPN is only designed to be a replacement for the RPN and does not have the Region of Interest (RoI) pooling, classification, and regression components. Therefore, using GPN alone cannot complete the ellipse detection pipeline. Moreover, we improve the KL divergence loss by using the Wasserstein distance.

\section{Data Preparation}
\label{sec:dataprep}
In this section, we discuss the procedures for preparing the lumber knot dataset. Specifically, Section~\ref{subsec:fixing} introduces the proposed algorithm to fix the misalignment in the raw images, while Section~\ref{subsec:knot_data} gives a brief introduction to the annotated lumber knot dataset.

\subsection{Preprocessing Algorithm}
\label{subsec:fixing}
Similar to the lumber image samples in Figure~\ref{fig: knot_example}, we have a total of $113$ pieces of lumber with four sides being scanned. The raw data do not contain any labels for knot faces, and the quality of the images is not ideal for object detection purposes. For example, it can be seen from Figure~\ref{fig: knot_example} that for each sample lumber image, there is one edge that is much longer than the other edge. Moreover, a major issue with the data is the pixel misalignment in the images of narrow surfaces. 

\begin{figure}[!ht]
	\centering
	\includegraphics[width= 0.47\textwidth]{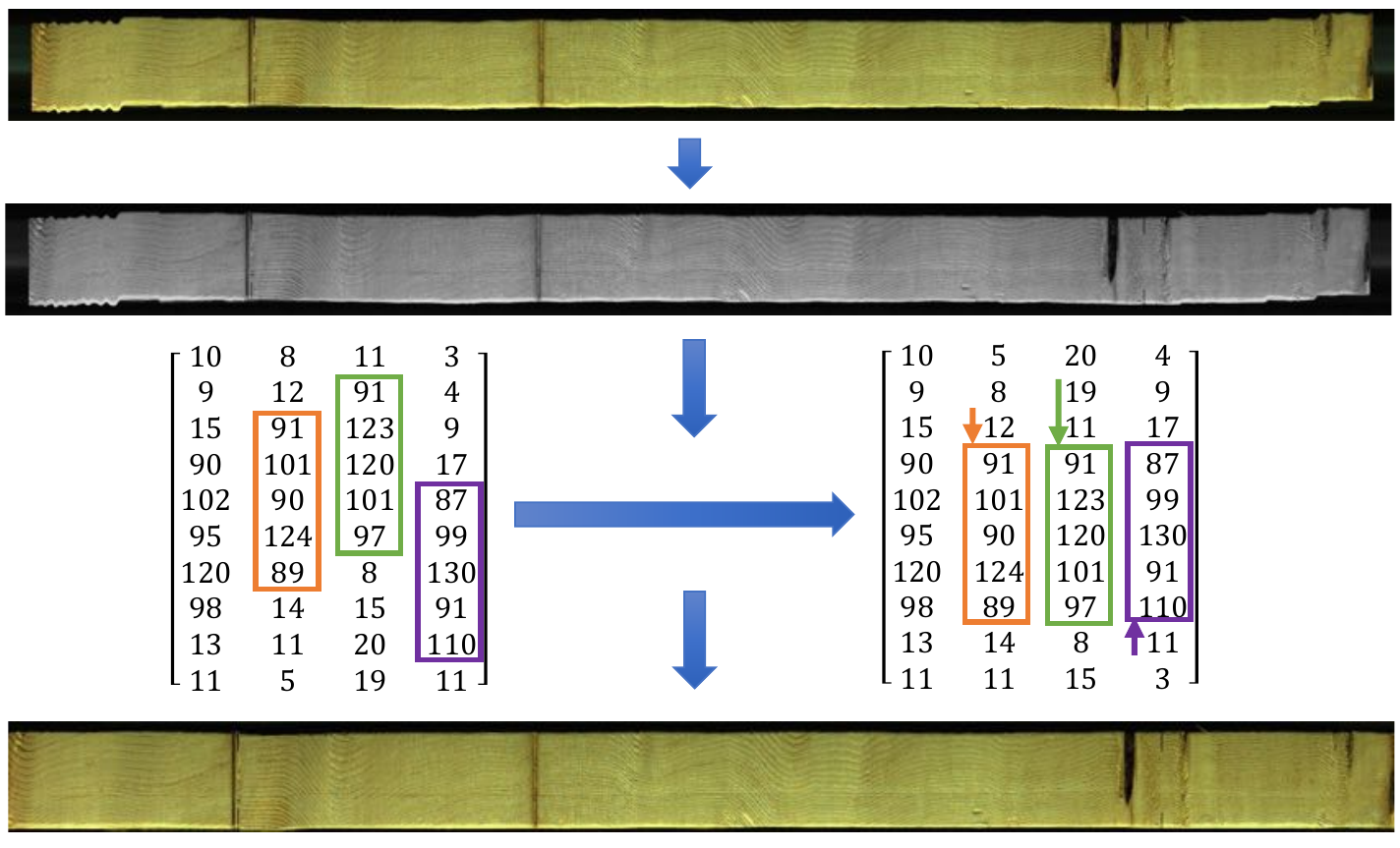}
	\caption{Illustration of fixing misalignment for one piece of lumber board. The RGB image is first converted to greyscale and fed to our algorithm to get the optimal displacement, which is used to shift the RGB image column-by-column to get the ideal result.}
	\label{fig:process}
\end{figure}

An example of pixel misalignment in the image data is illustrated in Figure~\ref{fig:process}. To effectively improve the data quality, the preprocessing step needs to simultaneously recognize and address two challenges: 
\begin{itemize}
	\item The backgrounds do not have uniformly black color and contain noise pixels. Specifically, the dark background color exhibits an arbitrary change pattern close to the lumber edges.
	
	\item The color of knot faces can be very similar to the background. If a knot lies on either edge of a lumber board, it is often difficult to distinguish it from the background purely based on its color.
\end{itemize}

These features make it impractical to fix the misalignment issue by simply setting up a threshold to distinguish the background from the lumber area. Preliminary analysis shows that methods based on thresholds produce non-smooth edges of the lumber region and may fail when the knots lie on either edge of the surface.

\begin{figure}[!ht]
	\centering
	\includegraphics[width= 0.47\textwidth]{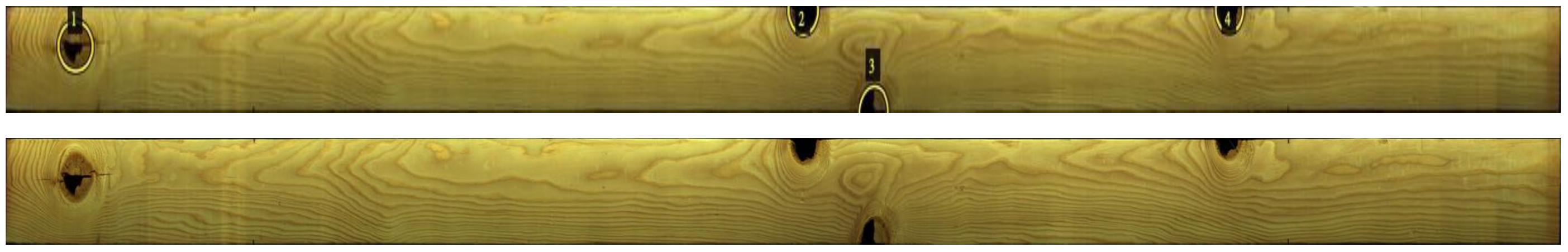}
	\caption{Bounding ellipse annotations for the visible knots on the wide surface of one sawn lumber.}
	\label{fig:labelexample}
\end{figure}

Taking these considerations into account, we propose an iterative algorithm to fix the misalignment problem through column-by-column alignment of pixels. We illustrate the procedures in Figure~\ref{fig:process}. An RGB image is first converted to grey scale. For each pixel column, the algorithm determines its optimal shift by comparing the current column with its previous shifted neighbouring columns by taking the sum of the vector norms that are weighted inversely by the distances between columns. That is, the weights are proportional to the inverse of the distance raised to the power $p$. Denote the pixel values in the $i$th column by $\boldsymbol{c}_i$. Let the number of neighbours be $n$ and the norm order be $k$. The optimal shift $\widehat{s}$ for column $i$, $i > 1$, is obtained by searching among all possible shifts $s\in \{s_{\min}^i, \dots, -1, 0, 1, \dots, s_{\max}^i\}$ such that
\begin{equation}
\widehat{s} = \argmin_{s\in\{s_{\min}^i, \dots, -1, 0, 1, \dots, s_{\max}^i\}} \sum_{j = \min(1, i - n)}^{i - 1}\frac{1}{|i-j|^p}||\boldsymbol{c}_i^{s} - \boldsymbol{c}_j||_k,
\end{equation}
where $\boldsymbol{c}_i^s$ represents pixels of the $i$th column shifted by $s$.

The values of $n$, $p$, and $k$ are fine-tuned to achieve optimal results. Taking the extent of misalignment into account, we set $n = 100$, $k = 2$, and $p = 1$ to search for the optimal shifts for each pixel column. This method proves to be effective in resolving the misalignment in the lumber images. A simplified example is depicted in Figure~\ref{fig:process}. It can be seen that our proposed greedy method is robust to the nonuniform dark background and edge-lying knots.

\subsection{Lumber Knot Dataset}
\label{subsec:knot_data}
After removing the dark background, lumber images are manually annotated using the free image annotation VGG Image Annotator \cite{dutta2019vgg} with each ellipse parameterized by five parameters: the $x$ and $y$ coordinates of the center of each ellipse denoted by $cx$ and $cy$; the semi-diameters along the $x$- and $y$-axes denoted by $rx$ and $ry$; the counterclockwise rotation angle in radians denoted by $\theta$.

Figure~\ref{fig:labelexample} shows an example of the bounding ellipses for all the visible knots on the wide surface of a piece of lumber. It can be seen that all the visible knots on the board are accurately and tightly annotated. We then randomly crop and resize the annotated images to generate square images for the detection task. This step also re-computes the parameters of the bounding ellipse based on their relative positions in each cropped image. The dimension of each cropped and resized image is 512$\times$512 pixels. For each piece of lumber, an average of $57.4$ cropped images that partially contain at least one elliptical knot are generated. The complete dataset contains 4894 annotated lumber knots and is freely accessible on \url{https://forestat.stat.ubc.ca/tiki-index.php?page=Databases+-+Products} to the public for future research. 

\section{Proposed Method}
\label{sec:approach}
In this section, we formulate our elliptical detection and localization method.
The input to our method is a single image. The outputs are detected ellipses parameterized by $(cx, cy, rx, ry, \theta)$.
Since we are interested in a single class of knots, there is no need to further classify the category to which the objects contained in the bounding ellipses belong.
Our approach combines Faster R-CNN and its region proposal networks with the GPN introduced in \cite{li2019detecting}. The overview of the model pipeline is discussed in Section~\ref{subsec:overview}.  We introduce the GPN in Section~\ref{subsec:gpn}. Our extensions in terms of the region proposal branch and ellipse regression loss functions are explained in Sections~\ref{subsec:region_prop}~and~\ref{subsec:alterloss}, respectively.

\subsection{Overview of the Model Pipeline}
\label{subsec:overview}

\begin{figure}[!ht]
	\centering
	\includegraphics[width= .45\textwidth]{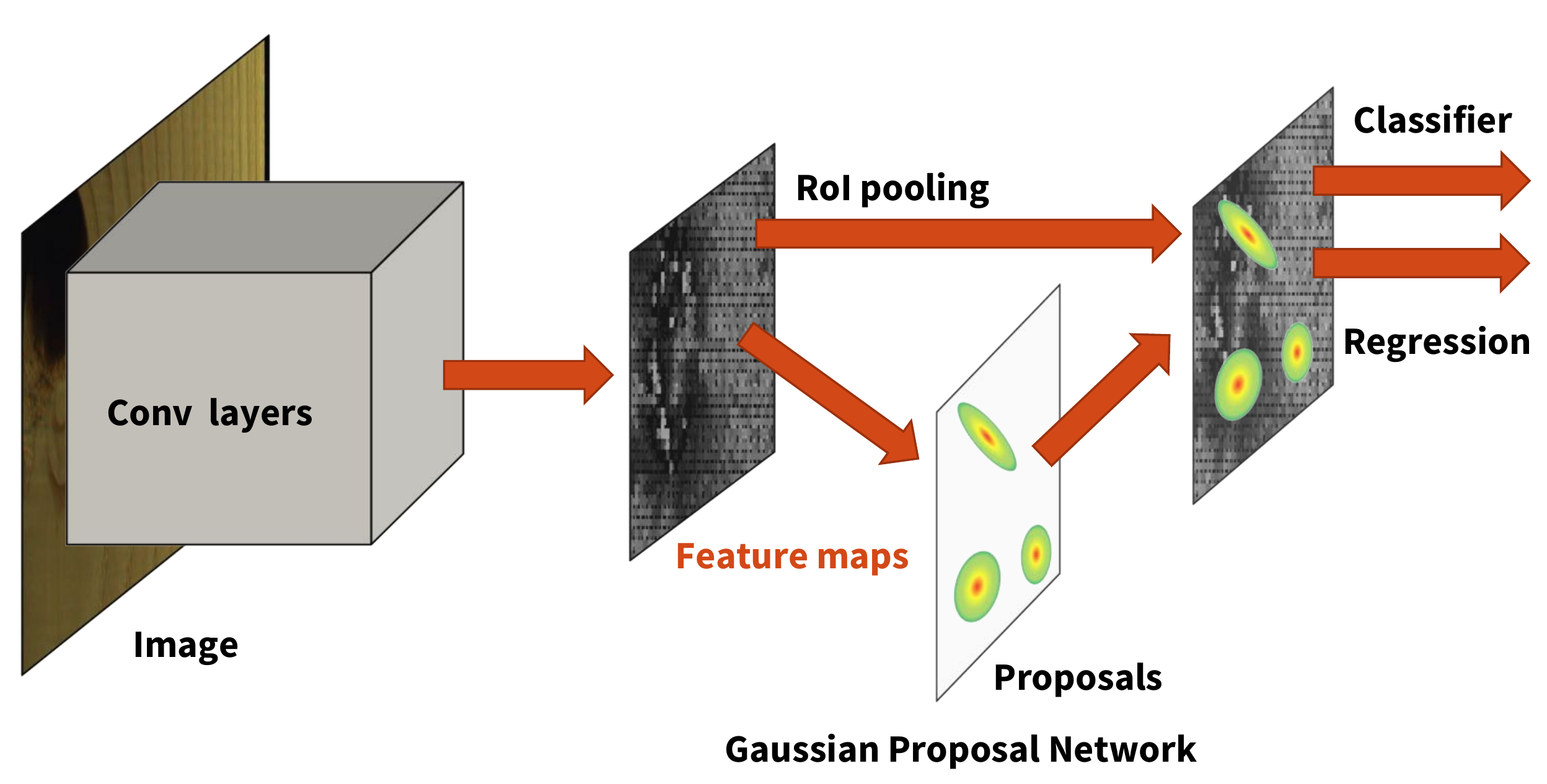}
	\caption{Overview of the ellipse localization and prediction model pipeline.}
	\label{fig:pip}
\end{figure}

As an illustration, Figure~\ref{fig:pip} shows the architecture of our proposed model. We adopt the basic architecture of the Faster R-CNN, which was originally designed to solve rectangular object detection problems. An image is first passed to the convolutional layers for feature map extraction. Instead of proposing bounding boxes, we use the GPN to propose bounding ellipses as 2D equi-probability contours of Gaussian distributions on the image plane. With the RoI pooling branch, the feature maps corresponding to the proposed regions are then obtained. The feature maps and proposals are finally fed to the ellipse regression and classification branch for final ellipse prediction.

\subsection{Gaussian Proposal Network}
\label{subsec:gpn}

In this section, we introduce the adaptations made by the GPN to accommodate Faster R-CNN for localizing ellipses, as required for the lumber knot detection problem. 

\subsubsection{Parameterizing Ellipses by Gaussian Distributions}
In \cite{li2019detecting}, ellipses are reparameterized as 2-dimensional contours of Gaussian distributions. As a result, the usual objective function for minimizing the L1 or L2 loss used in performing bounding box regression can be replaced by minimizing the distance metrics between two Gaussian distributions for ellipse regression. This section shows how ellipses can be represented using 2D Gaussian distributions.

An ellipse in a 2D coordinate system without rotation can be represented by
\begin{equation}
\frac{(x-\mu_x)^2}{\sigma_x^2} + \frac{(y-\mu_y)^2}{\sigma_y^2} = 1, \label{eq: ellipse_2d}
\end{equation}
where $\mu_x$ and $\mu_y$ are coordinates representing the centers of the ellipse, and $\sigma_x$ and $\sigma_y$ are the lengths of the semi-axes along the $x$ and $y$ axes. 

The probability density function of a 2D Gaussian distribution is given by
\begin{equation}
f(\boldsymbol{x}|\boldsymbol{\mu},\boldsymbol{\Sigma}) = \frac{\exp\left(-\frac{1}{2}(\boldsymbol{x}-\boldsymbol{\mu})^T \boldsymbol{\Sigma}^{-1} (\boldsymbol{x}-\boldsymbol{\mu}) \right)}{2\pi|\boldsymbol{\Sigma}|^{\frac{1}{2}}},
\end{equation}
where the vector $\boldsymbol{\mu}$ denotes the coordinate vector representing the center $(x, y)$, while $\boldsymbol{\mu}$ and $\boldsymbol{\Sigma}$ are the mean vector and covariance matrix of the Gaussian distribution. If we assume the off-diagonal entries of $\boldsymbol{\Sigma}$ are 0 and parameterize $\boldsymbol{\mu}$ and $\boldsymbol{\Sigma}$ as
\begin{equation}
\boldsymbol{\mu} =   
\begin{bmatrix}
\mu_x \\
\mu_y
\end{bmatrix}
\text{ and }
\boldsymbol{\Sigma} =   
\begin{bmatrix}
\sigma_x^2 & 0 \\
0 & \sigma_y^2
\end{bmatrix},
\end{equation}
the equation in Equation~\eqref{eq: ellipse_2d} for the ellipse corresponds to the density contour of the 2D Gaussian distribution when
\begin{equation}
(\boldsymbol{x}-\boldsymbol{\mu})^T \boldsymbol{\Sigma}^{-1} (\boldsymbol{x}-\boldsymbol{\mu}) = 1.
\end{equation}
When the major axis of the ellipse is rotated of an angle $\theta$ with respect to the $x$-axis, a rotation matrix $R(\theta)$ can be defined as
\begin{equation}
R(\theta) = 
\begin{bmatrix}
\cos\theta & \sin\theta \\
-\sin\theta & \cos\theta
\end{bmatrix}.
\end{equation}
This matrix can be used to map the coordinates in the original $(x, y)$ system into a new $(x', y')$ system, i.e.,
\begin{equation}
\begin{bmatrix}
x' \\
y'
\end{bmatrix} =
R(\theta)
\begin{bmatrix}
x \\
y
\end{bmatrix}
\end{equation}

Denote the lengths of the semi-major and semi-minor axes of the ellipse by $\sigma_l$ and $\sigma_s$. It can be shown that the ellipse centered at $(\mu_x, \mu_y)$ with semi-major and semi-minor axes of lengths $\sigma_l$ and $\sigma_s$, and a rotation angle of $\theta$ between its major axis and the $x$-axis where $\theta\in[-\frac{\pi}{2}, \frac{\pi}{2}]$ can be parameterized by a 2D Gaussian distribution with
\begin{equation}
\boldsymbol{\mu} =   
\begin{bmatrix}
\mu_x \\
\mu_y
\end{bmatrix}
\text{ and }
\boldsymbol{\Sigma}^{-1} = R^T(\theta)
\begin{bmatrix}
1/\sigma_l^2 & 0 \\
0 & 1/\sigma_s^2
\end{bmatrix}
R(\theta).\label{eq:cov}
\end{equation}

\subsubsection{Replacing Region Proposal Network with Gaussian Proposal Network}
The goal of the GPN is to propose bounding ellipses such that the Gaussian parameters $(\mu_x, \mu_y, \sigma_l, \sigma_s, \theta)$ are close to the ground truth ellipses through a distance metric. In \cite{li2019detecting}, Kullback-Leibler (KL) divergence is used as the distance measure. It is easily seen that the KL divergence between a proposed 2D Gaussian distribution $\mathcal{N}_p(\boldsymbol{\mu}_p, \boldsymbol{\Sigma}_p)$ and a target 2D Gaussian distribution $\mathcal{N}_t(\boldsymbol{\mu}_t, \boldsymbol{\Sigma}_t)$ is
\begin{align}
D_{KL}(\mathcal{N}_p || \mathcal{N}_t) &= \frac{1}{2}[\text{tr}(\boldsymbol{\Sigma}_t^{-1}\boldsymbol{\Sigma}_p) + (\boldsymbol{\mu}_p-\boldsymbol{\mu}_t)^T\boldsymbol{\Sigma}_t^{-1}(\boldsymbol{\mu}_p-\boldsymbol{\mu}_t) \nonumber\\
&\quad + \log\frac{|\boldsymbol{\Sigma}_t|}{|\boldsymbol{\Sigma}_p|} - 2],
\end{align}
where tr$(\cdot)$ is the trace of a matrix. GPN replaces RPN in the Faster R-CNN framework by predicting five parameters for ellipses instead of four parameters for bounding boxes and minimizing KL divergence instead of the smooth L1 loss. With these modifications, the RPN module can be replaced by the GPN to propose bounding ellipses.

\subsection{Region Proposal and Offset Regression}
\label{subsec:region_prop}

GPN was originally only designed to replace the RPN in the Faster R-CNN framework to generate elliptical proposals with the remaining components removed. Further detecting and predicting the exact locations of the ellipses in an image is out of the scope of GPN.
We re-introduce the necessary RoI pooling as well as the ellipse classification and regression components to complete the ellipse detection pipeline in analogy to Faster R-CNN. Faster R-CNN pools the feature maps in a rectangular region around the detection to have a fixed resolution, applies a CNN to these, and predicts offsets between the proposed parameters and the ground-truth parameters for each detection using a fully-connected layer. Similarly, given each ellipse proposal output by the GPN, we find the tightest axis-aligned bounding box covering the proposal. Feature maps in this rectangular are pooled in a similar manner to the Faster R-CNN. A fully-connected layer is then used to predict the parameter offsets based on the proposal center as well as the ellipse direction and shape for each detected ellipse. 

\subsection{Loss Function for Ellipse Regression}
\label{subsec:alterloss}
In Faster R-CNN, L1 or L2 losses are used as the loss function to predict the offsets between the four pairs of predicted and ground-truth parameters defining a rectangular object. However, these losses do not work well for elliptical object detection problems since the angle parameter needs to be treated differently than the other four parameters.

A natural choice for the ellipse regression loss is the KL divergence in GPN, which is used as the distance measure between two Gaussian distributions in generating ellipse proposals. Nevertheless, KL divergence has a few non-negligible drawbacks. Firstly, KL divergence is asymmetric, i.e., $D_{KL}(\mathcal{D}_1||\mathcal{D}_2)$ does not always equal $D_{KL}(\mathcal{D}_2||\mathcal{D}_1)$ given two distributions $\mathcal{D}_1$ and $\mathcal{D}_2$. Secondly, KL divergence can sometimes be numerically unstable and it tends to be very large in magnitude when the two distributions are far apart. This may cause problems in gradient evaluation, hindering convergence of the neural network. Lastly, KL divergence does not satisfy the triangle inequality and is not a valid mathematical metric in that sense.

Wasserstein distance is another popular distance measure defined between two probability distributions. In contrast to KL divergence, Wasserstein distance is symmetric and satisfies the triangle inequality. In recent years, Wasserstein distance has been proposed to replace other asymmetric losses to improve the results generated by neural network models. For example, Arjovsky \textit{et al.} \cite{arjovsky2017wasserstein} propose Wasserstein GAN to stabilize the training of GANs. 

The $p$-Wasserstein distance between two probability measures $\mu$ and $\nu$ is
\begin{equation}
W_p(\mu, \nu) = \left(\inf \mathbb{E} [d(X,Y)^p]\right)^{1/p},
\end{equation}
where $d(\cdot, \cdot)$ is a norm function, $X\sim\mu$, and $Y\sim\nu$. For two 2D Gaussian distributions, the 2-Wasserstein distance between them with respect to the usual Euclidean norm has a convenient form. For a proposed 2D Gaussian distribution $\mathcal{N}_p(\boldsymbol{\mu}_p, \boldsymbol{\Sigma}_p)$ and a target 2D Gaussian distribution $\mathcal{N}_t(\boldsymbol{\mu}_t, \boldsymbol{\Sigma}_t)$, the 2-Wasserstein distance with respect to the Euclidean norm is
\begin{align}
[W_2(\mathcal{N}_p, \mathcal{N}_t)]^2 &= ||\boldsymbol{\mu}_p-\boldsymbol{\mu}_t||^2_2 +\text{tr}\Big[\boldsymbol{\Sigma}_p + \boldsymbol{\Sigma}_t \nonumber\\
&\quad- 2\left(\boldsymbol{\Sigma}_p^{\frac{1}{2}}\boldsymbol{\Sigma}_t\boldsymbol{\Sigma}_p^{\frac{1}{2}}\right)^{\frac{1}{2}}\Big]
\end{align}
according to the results in \cite{olkin1982distance}, where $\text{tr}(\cdot)$ is the trace of a matrix. 
In the commutative case where $\boldsymbol{\Sigma}_p \boldsymbol{\Sigma}_t = \boldsymbol{\Sigma}_t \boldsymbol{\Sigma}_p$, the 2-Wasserstein distance between two 2D Gaussian distributions can be further simplied to 
\begin{equation}
[W_2(\mathcal{N}_p, \mathcal{N}_t)]^2 = ||\boldsymbol{\mu}_p-\boldsymbol{\mu}_t||^2_2 + ||\boldsymbol{\Sigma}_p^{\frac{1}{2}}-\boldsymbol{\Sigma}_t^{\frac{1}{2}}||^2_F,
\end{equation}
where $||\cdot||_F$ is the Frobenius norm of a matrix. The two covariance matrices can be computed based on the inverses in Equation~\eqref{eq:cov} from the ellipse parameters. The square root matrices $\boldsymbol{\Sigma}_p^{\frac{1}{2}}$ and $\boldsymbol{\Sigma}_t^{\frac{1}{2}}$ have a closed-form solution:
\begin{equation}
\boldsymbol{\Sigma}_p^{\frac{1}{2}} = R^T_p(\theta)
\begin{bmatrix}
\sigma_{p,l} & 0 \\
0 & \sigma_{p,s}
\end{bmatrix}
R_p(\theta)
\end{equation}
and
\begin{equation}
\boldsymbol{\Sigma}_t^{\frac{1}{2}} = R^T_t(\theta)
\begin{bmatrix}
\sigma_{t,l} & 0 \\
0 & \sigma_{t,s}
\end{bmatrix}
R_t(\theta).
\end{equation}

With all these modifications and adaptions, the overall loss function is the weighted sum of three components: the GPN ellipse proposal loss, the ellipse regression loss, and the cross entropy of classifying ellipses and backgrounds.

\section{Experimental Results}
\label{sec:exp}

Comprehensive evaluations of our proposed method for detecting elliptical knots in the lumber knot dataset are presented in this section. We introduce the experimental setup in Section~\ref{subsec:exp_setup} and summarize the quantitative detection performances across different experiment settings in Section~\ref{subsec:detection_quan}. Visualizations of the detection results are provided in Section~\ref{subsec:detection_qual}. We also compare our proposed method against the baseline deep learning-based solution and geometric reconstructions in~\ref{subsec:comp}.

\subsection{Experiment Setup}
\label{subsec:exp_setup}
Among all the annotated lumber specimens in the lumber knot dataset, 70\% are randomly chosen as the training set, 10\% are chosen as the validation set, and the remaining 20\% are used as the test set. We trained the model on the training set for 20 epochs; the model with the lowest validation total loss is saved for testing purposes. As with the original Faster R-CNN model, the pretrained VGG-16 network in \cite{simonyan2014very} is used as the base model for ellipse proposal generation and feature map extraction. We trained our proposed model with PyTorch 1.0. 

The average of the intersection over union (IoU) between all the detected ellipses and ground truth ellipses are computed and used as the metrics to evaluate the performance of a proposed detection methods. Since there are not closed-formed formula to compute the overlapping area between two ellipses, we draw a grid of points from both ellipses and use a discretized sampling method to compute the IoU between the two ellipses. The width and height of the point grid are the same as those of the tightest axis-aligned rectangle covering both ellipses. One point is assigned at each pixel location.

\subsection{Quantitative Evaluation of Detection Performance}
\label{subsec:detection_quan}
To evaluate the performance of our proposed method, four experimental settings are considered as follows:
\begin{enumerate}
	\item \textit{RPN with L2 loss for ellipse offset regression.} RPN in the Faster R-CNN is used to generate bounding ellipse proposals. Instead of the four parameters characterizing a rectangle, the RPN outputs the five parameters characterizing an ellipse. L2 loss is used for the ellipse offset regression. This setting directly modifies the RPN for the ellipse detection problem and serves as the baseline model.
	\item \textit{GPN with L2 loss for ellipse offset regression.} RPN in the Faster R-CNN is replaced with the GPN to generate ellipse proposals. L2 loss is used for ellipse offset regression.
	\item \textit{GPN with KL divergence for ellipse offset regression.}
	\item \textit{GPN with 2-Wasserstein distance for ellipse offset regression.} 
\end{enumerate}

Note that GPN uses KL divergence while RPN uses smooth L1 loss in the proposal network. We trained the model five times under each setting. The IoUs between the detected and ground-truth ellipses under each setting in each repeated experiment along with the mean IoU and standard error are reported in Table~\ref{tab: exp}.

\setlength{\tabcolsep}{4pt}
\begin{table}
	\centering
	\begin{tabular}{*{5}{l}}
		\noalign{\smallskip}
		\toprule
		Setting & RPN, L2 & GPN, L2 & GPN, KLD & GPN, 2-WD \\\midrule
		Exp 1 & 60.16 & 64.74 & 73.37 & 73.00 \\
		Exp 2 & 65.59 & 66.14 & 73.13 & 73.32 \\
		Exp 3 & 64.54 & 65.82 & 72.48 & 72.98 \\
		Exp 4 & 65.98 & 66.97 & 71.07 & 72.85 \\
		Exp 5 & 61.87 & 65.40 & 73.26 & 73.10 \\\midrule
		Mean IoU & 63.63 & 65.81 & 72.66 & 73.05 \\
		s.e. & 2.52 & 0.83 & 0.95 & 0.18\\
		\bottomrule
	\end{tabular}
	
	\caption{The IoUs between the detected and ground-truth ellipses under each of the four settings in each repeated experiment along with the mean IoU and the standard error. L2, KLD, and 2-WD represent L2 loss, KL divergence, and 2-Wasserstein distance, respectively. All numbers are in percentages.
	}
	\label{tab: exp}
\end{table}
\setlength{\tabcolsep}{1.4pt}


From Table~\ref{tab: exp}, it can be seen that using our proposed method, which generates elliptical proposals with GPN and uses the 2-Wasserstein distance as the loss function for ellipse offeset regression, improves the mean IoU by 10\% compared with a general detector such as Faster R-CNN, which generates proposals with RPN and uses L2 distance as the loss function for offset regression. Directly replacing RPN in Faster R-CNN with GPN improves the mean IoU by around 2\%. Replacing the general-purpose L2 loss with KL divergence or 2-Wasserstein distance both significantly improves the ellipse detection performance. In particular, models using 2-Wasserstein distance outperform models using KL divergence by more than 0.4\%. Moreover, using the Wasserstein distance instead of KL divergence led to a much lower (0.18 vs. 0.95) standard deviation across experiments.

\subsection{Qualitative Evaluation of Detection Performance}
\label{subsec:detection_qual}
In the previous section, we quantitatively evaluated the performance of our proposed ellipse detection method against the baseline Faster R-CNN model. In this section, visualizations of the detected ellipses in the lumber knot dataset are provided to qualitatively assess our proposed method. 

\begin{figure}[!ht]
	\centering
	\begin{subfigure}{.115\textwidth}
		\centering
		\includegraphics[width=0.95\linewidth]{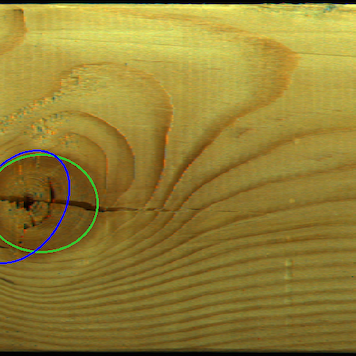}
	\end{subfigure}
	\begin{subfigure}{.115\textwidth}
		\centering
		\includegraphics[width=0.95\linewidth]{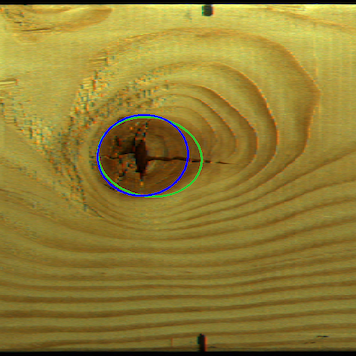}
	\end{subfigure}
	\begin{subfigure}{.115\textwidth}
		\centering
		\includegraphics[width=0.95\linewidth]{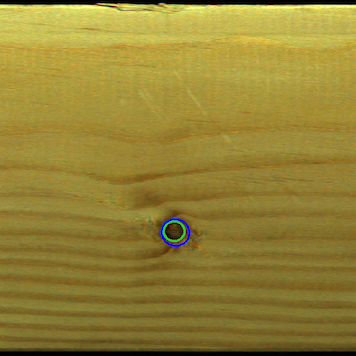}
	\end{subfigure}
	\begin{subfigure}{.115\textwidth}
		\centering
		\includegraphics[width=0.95\linewidth]{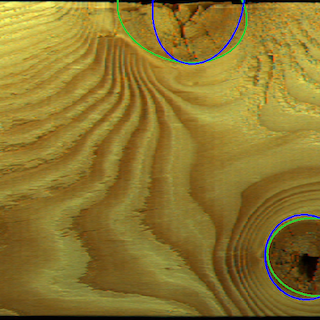}
	\end{subfigure}
	
	\begin{subfigure}{.115\textwidth}
		\centering
		\includegraphics[width=0.95\linewidth]{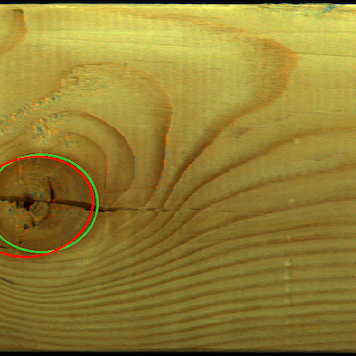}
	\end{subfigure}
	\begin{subfigure}{.115\textwidth}
		\centering
		\includegraphics[width=0.95\linewidth]{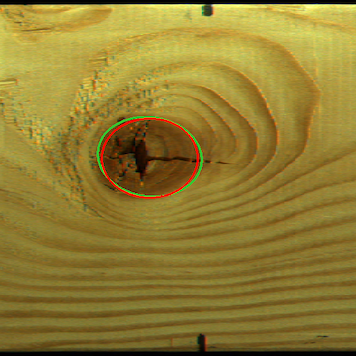}
	\end{subfigure}
	\begin{subfigure}{.115\textwidth}
		\centering
		\includegraphics[width=0.95\linewidth]{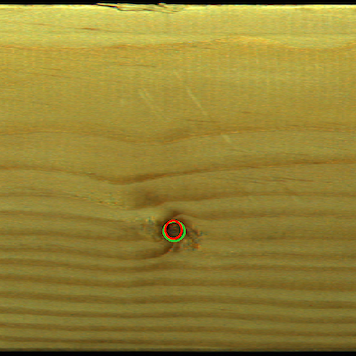}
	\end{subfigure}
	\begin{subfigure}{.115\textwidth}
		\centering
		\includegraphics[width=0.95\linewidth]{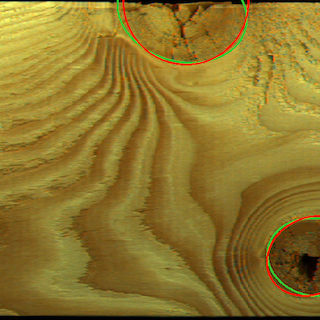}
	\end{subfigure}

	\begin{subfigure}{.115\textwidth}
		\centering
		\includegraphics[width=0.95\linewidth]{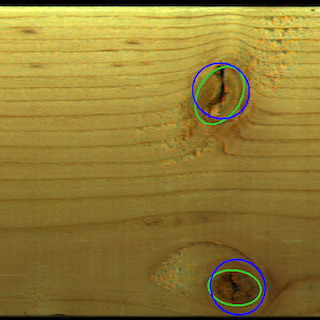}
	\end{subfigure}
	\begin{subfigure}{.115\textwidth}
		\centering
		\includegraphics[width=0.95\linewidth]{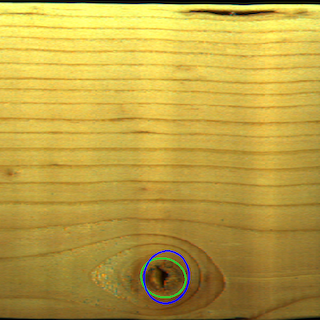}
	\end{subfigure}
	\begin{subfigure}{.115\textwidth}
		\centering
		\includegraphics[width=0.95\linewidth]{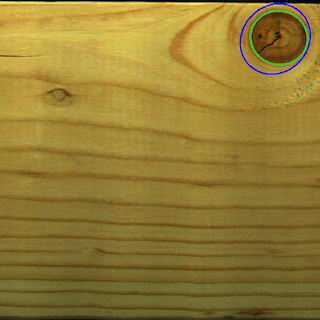}
	\end{subfigure}
	\begin{subfigure}{.115\textwidth}
		\centering
		\includegraphics[width=0.95\linewidth]{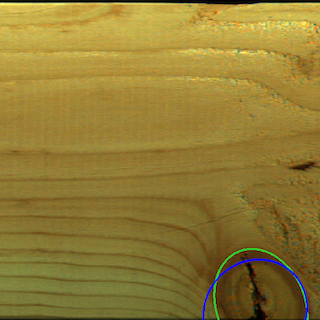}
	\end{subfigure}
	
	\begin{subfigure}{.115\textwidth}
		\centering
		\includegraphics[width=0.95\linewidth]{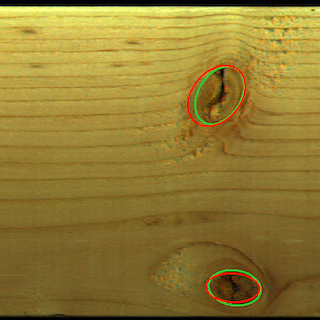}
	\end{subfigure}
	\begin{subfigure}{.115\textwidth}
		\centering
		\includegraphics[width=0.95\linewidth]{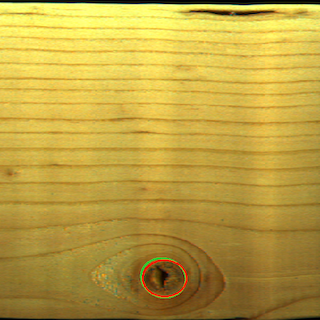}
	\end{subfigure}
	\begin{subfigure}{.115\textwidth}
		\centering
		\includegraphics[width=0.95\linewidth]{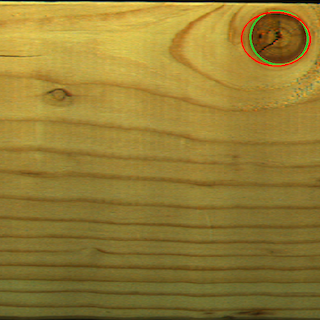}
	\end{subfigure}
	\begin{subfigure}{.115\textwidth}
		\centering
		\includegraphics[width=0.95\linewidth]{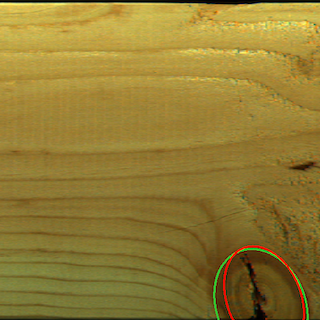}
	\end{subfigure}

	\begin{subfigure}{.115\textwidth}
		\centering
		\includegraphics[width=0.95\linewidth]{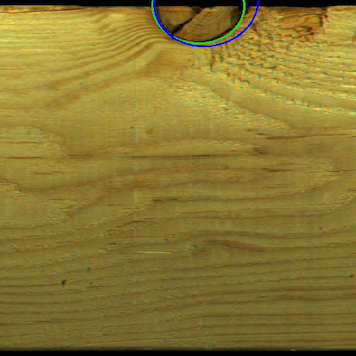}
	\end{subfigure}
	\begin{subfigure}{.115\textwidth}
		\centering
		\includegraphics[width=0.95\linewidth]{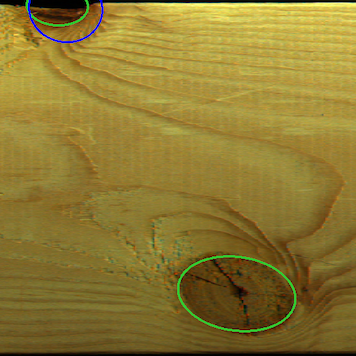}
	\end{subfigure}
	\begin{subfigure}{.115\textwidth}
		\centering
		\includegraphics[width=0.95\linewidth]{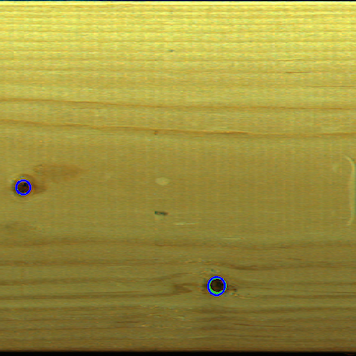}
	\end{subfigure}
	\begin{subfigure}{.115\textwidth}
		\centering
		\includegraphics[width=0.95\linewidth]{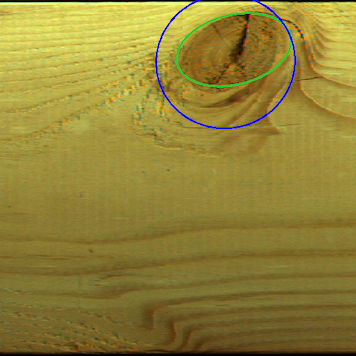}
	\end{subfigure}
	\begin{subfigure}{.115\textwidth}
		\centering
		\includegraphics[width=0.95\linewidth]{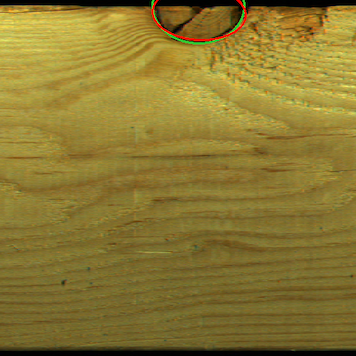}
	\end{subfigure}
	\begin{subfigure}{.115\textwidth}
		\centering
		\includegraphics[width=0.95\linewidth]{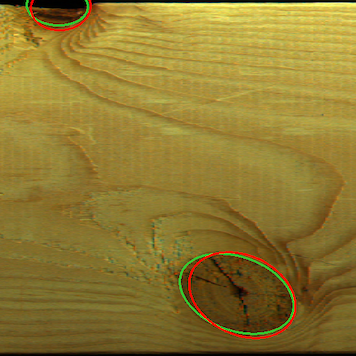}
	\end{subfigure}
	\begin{subfigure}{.115\textwidth}
		\centering
		\includegraphics[width=0.95\linewidth]{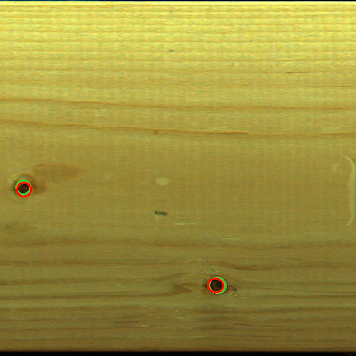}
	\end{subfigure}
	\begin{subfigure}{.115\textwidth}
		\centering
		\includegraphics[width=0.95\linewidth]{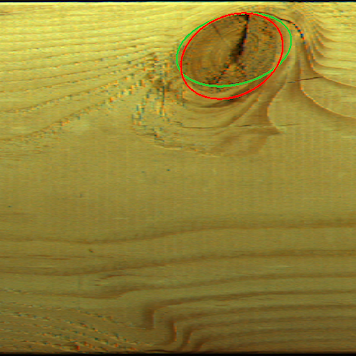}
	\end{subfigure}
	\caption{Examples of detected ellipses from three specimens of lumber using the baseline method (RPN, L2) and our proposed method (GPN, 2-WD). Images in every two rows are from the same specimen. Green, blue, and red ellipses are the ground-truth ellipses, the detected ellipses using the baseline method, and the detected ellipses using our proposed method, respectively.}
	\label{fig: detection}
\end{figure}

Figure~\ref{fig: detection} shows the ground-truth knots from three pieces of lumber as well as the detected knots  using the baseline method (RPN, L2) and our proposed method (GPN, 2-WD). It can be seen that the elliptical bounds for knots are drawn more tightly and accurately using our proposed method compared with the baseline method. This holds particularly true for knots close to the board boundary, which are only partially visible. Nonetheless, it can be seen that the rotation angles of the detected ellipses have relatively large errors in some of the examples shown in Figure~\ref{fig: detection}. Future research can be done to further improve the prediction performance for the rotation angles, possibly by introducing oriented instead of axis-aligned pooling regions in the GPN.

\subsection{Comparisons to Geometric Ellipse Detection} \label{subsec:comp}
\begin{figure}[!ht]
	\centering
	\begin{subfigure}{.091\textwidth}
		\centering
		\includegraphics[width=0.95\linewidth]{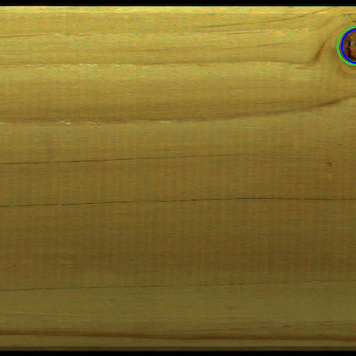}
	\end{subfigure}
	\begin{subfigure}{.091\textwidth}
		\centering
		\includegraphics[width=0.95\linewidth]{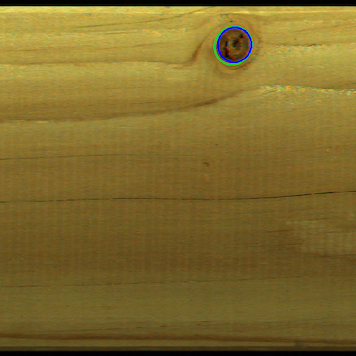}
	\end{subfigure}
	\begin{subfigure}{.091\textwidth}
		\centering
		\includegraphics[width=0.95\linewidth]{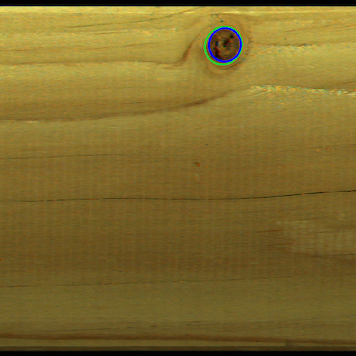}
	\end{subfigure}
	\begin{subfigure}{.091\textwidth}
		\centering
		\includegraphics[width=0.95\linewidth]{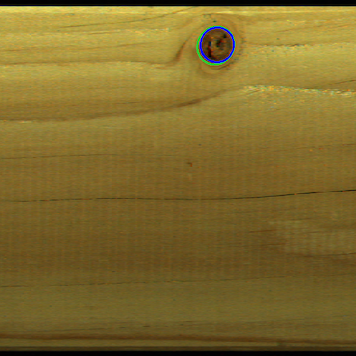}
	\end{subfigure}
	\begin{subfigure}{.091\textwidth}
		\centering
		\includegraphics[width=0.95\linewidth]{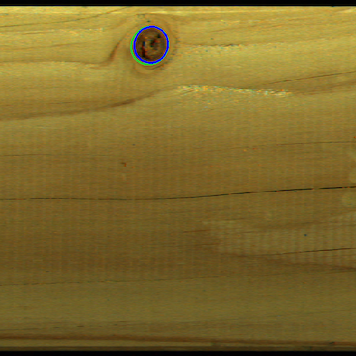}
	\end{subfigure}
	\begin{subfigure}{.091\textwidth}
		\centering
		\includegraphics[width=0.95\linewidth]{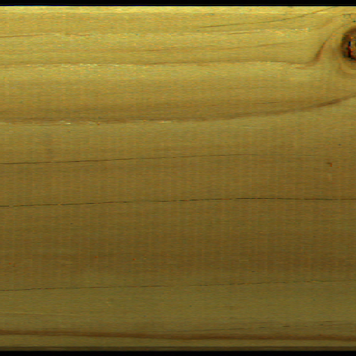}
	\end{subfigure}
	\begin{subfigure}{.091\textwidth}
		\centering
		\includegraphics[width=0.95\linewidth]{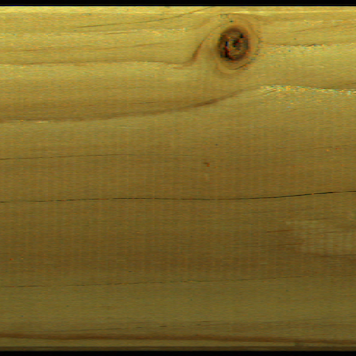}
	\end{subfigure}
	\begin{subfigure}{.091\textwidth}
		\centering
		\includegraphics[width=0.95\linewidth]{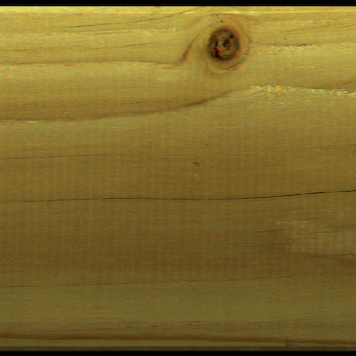}
	\end{subfigure}
	\begin{subfigure}{.091\textwidth}
		\centering
		\includegraphics[width=0.95\linewidth]{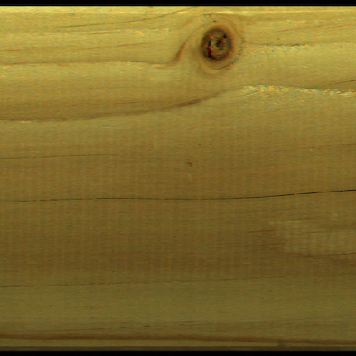}
	\end{subfigure}
	\begin{subfigure}{.091\textwidth}
		\centering
		\includegraphics[width=0.95\linewidth]{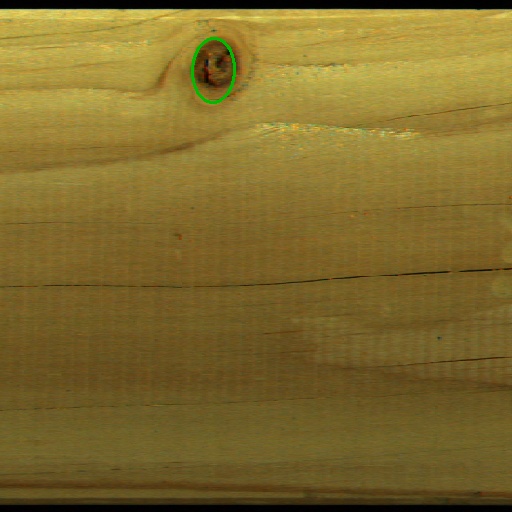}
	\end{subfigure}
	
	\caption{Examples of detecting the same ellipse in different cropped images using our method (GPN, 2-WD) and Jia's method. Green and blue ellipses in the first row are the ground-truth ellipses and the detected ellipses using our proposed method. Green ellipses in the second row are the detected ellipses using Jia's method. Note that Jia's method only detects the knot in the last cropped image.}
	\label{fig: detection2}
\end{figure}

To compare with our proposed method, we applied the geometric fast ellipse detector using projective invariant pruning method proposed in \cite{jia2017fast} to the lumber knot dataset (their code is available on \url{https://github.com/dlut-dimt/ellipse-detector}). Jia's method uses geometric features to find ellipses, which performs extremely poorly on lumber knot images. Among all the test images, less than 1\% of the ellipses can be detected using their method. In particular, Jia's method failed to detect any ellipses in all the 12 images in Figure~\ref{fig: detection}. Furthermore, Jia's method is also sensitive to the positioning of ellipses. For example, Figure~\ref{fig: detection2} visualizes the detection results using our proposed method (GPN + 2-WD) versus Jia's method on the same knot across different cropped images of a lumber board. It can be seen that among these five positions, our method consistently generates accurate position estimations while Jia's method is only able to successfully yet inaccurately detect this ellipse at one out of the five positions. Therefore, we conclude that our method is more robust and reliable than non-machine learning-based ellipse methods and works much better on detecting knots in lumber images. In terms of runtime, for an image of 512$\times$512 pixels,  Jia's method takes around 8.8 milliseconds to generate predictions, while our proposed method (GPN + 2-WD) takes around 226 milliseconds in the test stage. Although Jia's method is faster than our proposed method, both methods are sufficiently fast for real-world applications.

\section{Conclusion}
\label{sec:conc}
In this paper, we propose a method tailored to detect and localize elliptical objects in images. Our method adapts the Region Proposal Network in Faster R-CNN to model elliptical objects. We also extend the existing Gaussian Proposal Network by adding RoI pooling as well as the ellipse classification and regression branches to complete the elliptical object detection pipeline. Furthermore, we propose Wasserstein distance as the loss function for ellipse offset predictions. Experiments on the sawn lumber images show that our proposed method improves the mean IoU of the detected ellipses by 10\% compared with the baseline method. This is a major improvement in this illustrative application in so far as the result will be used in models that predict lumber strength and hence determine the grade class into which a piece of lumber will be placed. That in turn benefits consumers who use these grades in selecting lumber for their specific application. And it benefits producers who will use machine grading techniques to classify their lumber. For the forest products industry, this can result in products of better quality for intended use and in turn improvements to the manufacturer's bottom line. In addition, specific to the lumber example, we propose an algorithm to correct the misalignment of raw images of lumber specimens and create the first open-source lumber knot dataset with labeled elliptical knots, which can be used for future research.  While experiments in this paper focus on the knot detection problem in sawn lumber images, our proposed method can easily be applied to detect ellipses in datasets containing other types of elliptical objects. In future work, we will attempt to predict the 3D knot shapes given 2D ellipsoid supervisions from both sides. It would also be interesting to predict the entire elliptical cone given images of all three sides, supervised by the 2D ellipse/Gaussian formulation.


\noindent
\textbf{Acknowledgements.} The authors thank Conroy Lum, along with FPInnovations and its technical support staff, for facilitating the experimental work that was done to produce the data used in this paper. We also thank Terry Chen, Zongjun Liu, Angela Wang-Lin, and Caitlin Zhang for their assistance in data processing. The work reported in this paper was partially supported by FPInnovations and a Collaborative Research and Development grant from the Natural Sciences and Engineering Research Council of Canada. 

\section*{Appendix}
\appendix
\section{Equipment and Methods}

To demonstrate the methodology proposed in this paper, we scanned the lumber specimens in the wood products testing laboratory at FPInnovations. The lumber species used in this experimental study is Douglas Fir (DF) lumber, which is widely used to make dimensional lumber. The DF specimens are of dimension $2'' \times 6'' \times 12'$. The selected specimens represent a test material with adequate variation in appearance to ensure a good sample for studying different knot patterns, which could arise from different tree growth conditions and sawing practices. The total number of specimens used in this study was $113$.

\begin{figure}[!ht]
	\centering
	\includegraphics[width=0.47\textwidth]{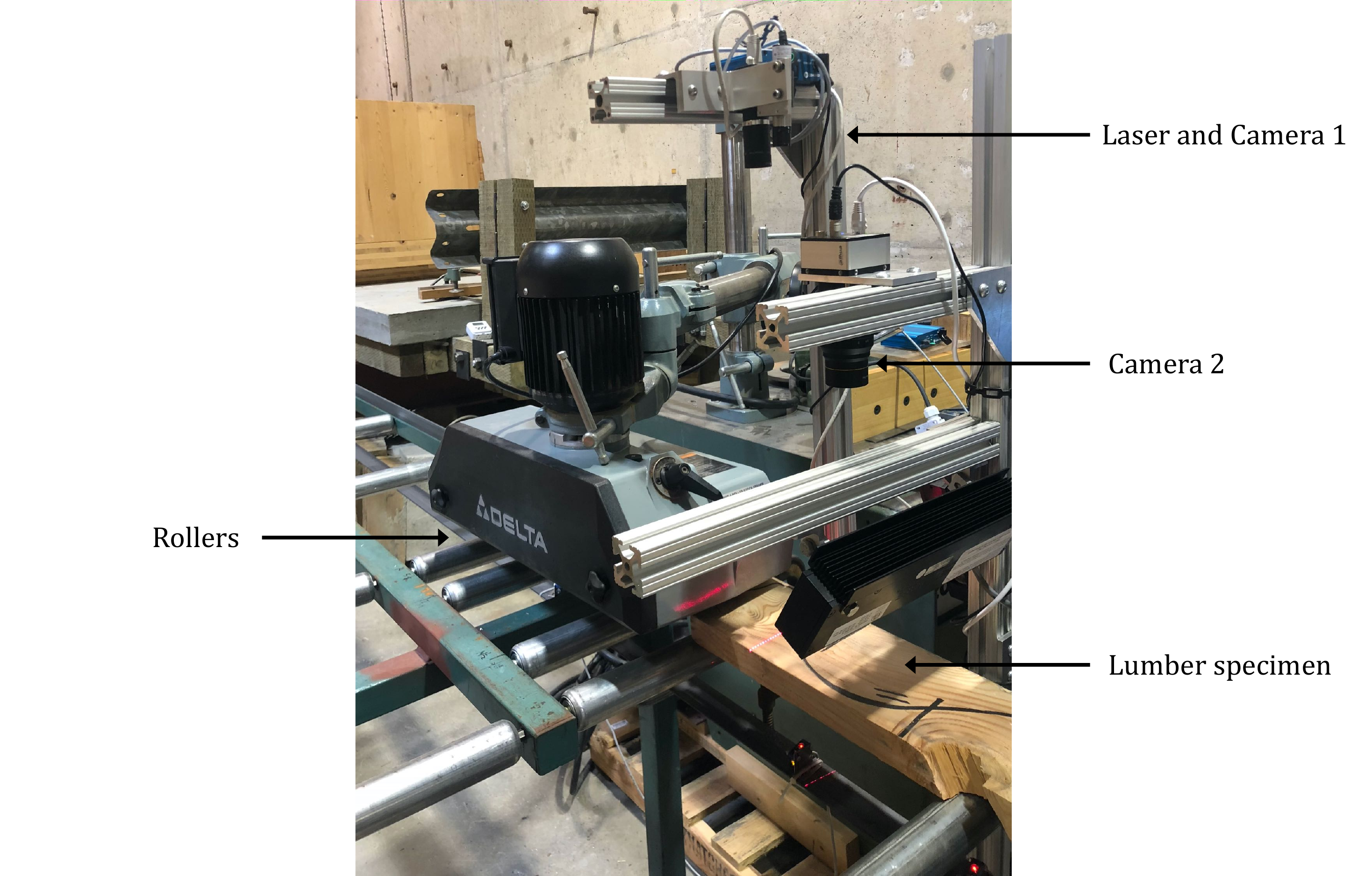}
	\caption{Setup of the lumber scanner. The lumber specimen can be seen at the lower right corner of the figure, on the rollers on which it moves it forward at a constant speed under the cameras.}
	\label{fig:tracheid_scanner}
\end{figure}

\begin{figure}[!ht]
	\centering
	\includegraphics[width=0.47\textwidth]{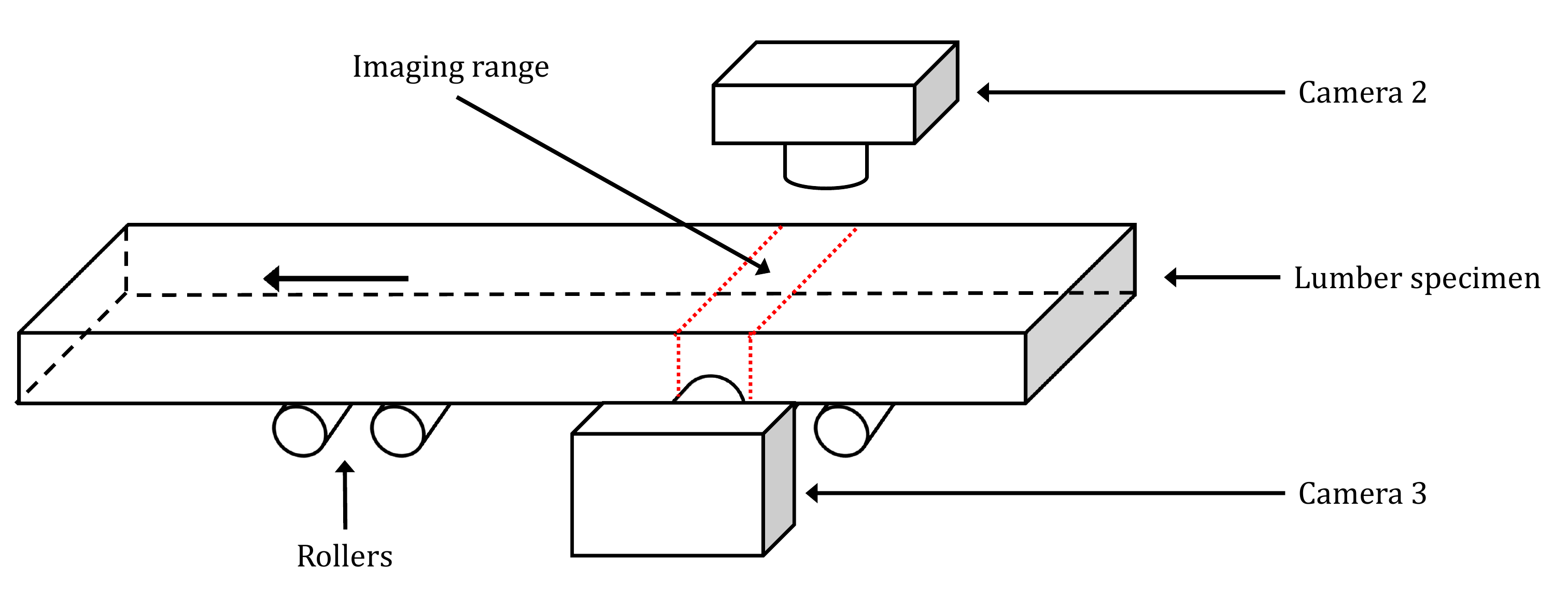}
	\caption{Schematic diagram of the color camera setup.}
	\label{fig:color_camera}
\end{figure}

As shown in Figure~\ref{fig:tracheid_scanner}, the lumber scanner is equipped with two sets of digital cameras. The set of cameras that take the color images of the lumber surfaces are Camera 2 and Camera 3, which, however, cannot be seen from Figure~\ref{fig:tracheid_scanner}. For illustrative purposes, a schematic setup of the color cameras is given in Figure~\ref{fig:color_camera}.

The issue of pixel misalignment results from the setting used in the lumber scanner, where each lumber specimen is sent to the camera through rollers at a fixed speed. The two color cameras frequently take images at the same time from both the top and side of the specimen. The scanner produces lumber images using its built-in image stitching methods. For each lumber specimen, four color images in PNG format are generated with a resolution of $8000\times2048$ pixels. Note that although each piece of lumber has six sides, the two ends are typically ignored since their area is much smaller than the other four sides. As natural materials, sawn lumber does not necessarily have perfectly flat surfaces. As a result, inevitable `bumpiness' is generated when a piece of lumber is scanned as the image stitching algorithm cannot automatically recognize random shifts of the lumber surface. This leads to the misalignment of pixels in the scanned output images, and a similar phenomenon can also occur when applying computer vision-based industrial inspection on other materials that are scanned on rollers. 

\section{Image Preprocessing: Comparisons with Method Based on Threshold}
Raw scanned images of lumber specimens exhibit different levels of distortions. 
Lumber specimens experience the most `bumping' at the beginning and end of the scanning process, often causing the most severe distortions to occur at the two ends of the scanned images for a piece of lumber. 

\begin{figure}[ht]
	\centering
	\begin{minipage}[b]{\linewidth}
		\begin{subfigure}{\textwidth}
			\centering
			\includegraphics[width=0.9\linewidth]{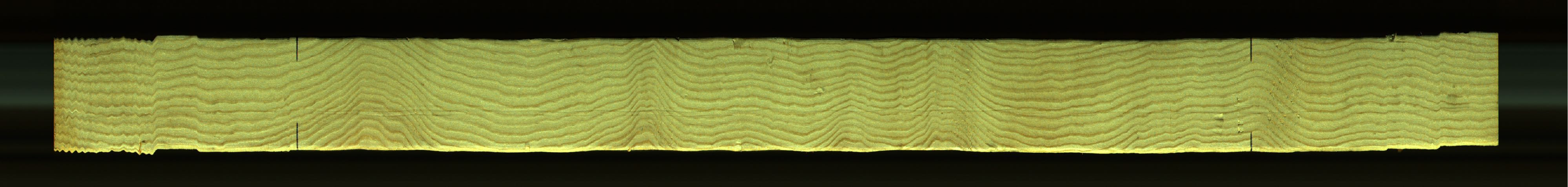}
		\end{subfigure}
		\subcaption{Moderate distortion with no or minor defects}\label{fig:1a}
		
		\begin{subfigure}{\textwidth}
			\centering
			\includegraphics[width=0.9\linewidth]{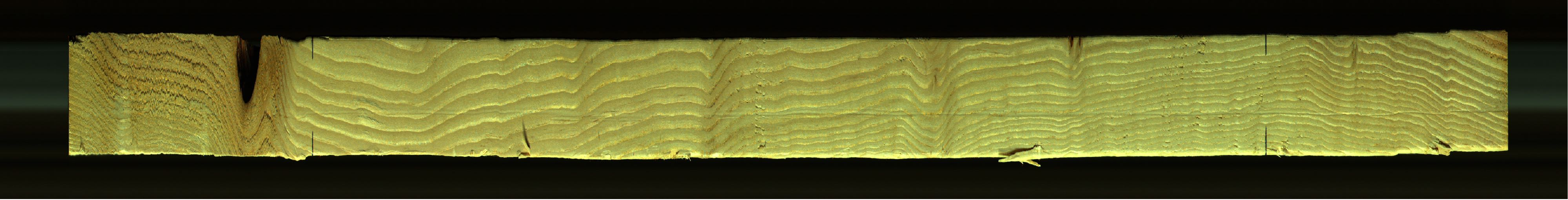}
		\end{subfigure}
		\subcaption{Moderate distortion with evident defects}\label{fig:1b}
		
		\begin{subfigure}{\textwidth}
			\centering
			\includegraphics[width=0.9\linewidth]{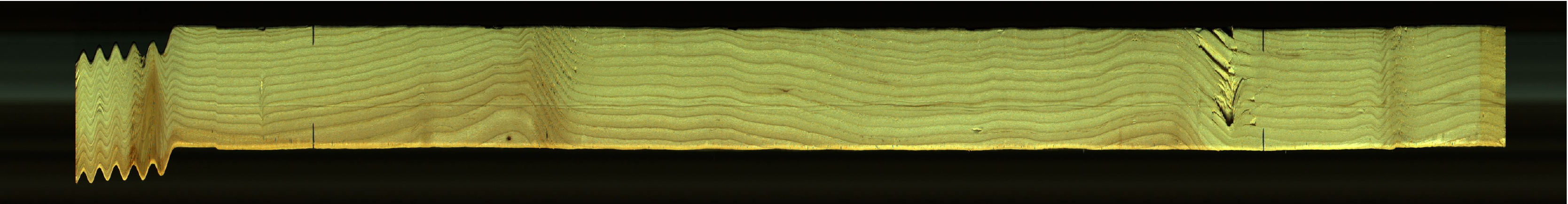}
		\end{subfigure}
		\subcaption{Severe distortion with no or minor defects}\label{fig:1c}
		
		\begin{subfigure}{\textwidth}
			\centering
			\includegraphics[width=0.9\linewidth]{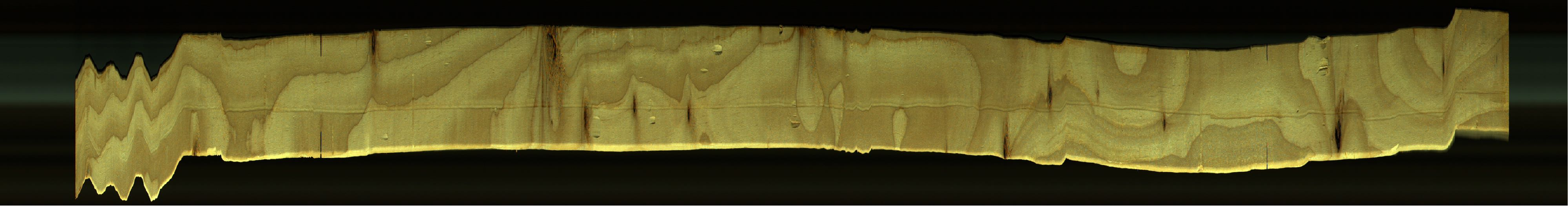}
		\end{subfigure}
		\subcaption{Severe distortion with evident defects}\label{fig:1d}
	\end{minipage}
	\caption{Examples of common distortion cases in the dataset.}
	\label{fig:examples}
\end{figure}

Figure~\ref{fig:examples} shows four most common cases of image distortions in the dataset. Figures~\ref{fig:1a} and~\ref{fig:1b} correspond to the cases when there are moderate distortions. Lumber images in these two figures show minor pixel misalignment at the two ends. The middle part of the lumber images bends slightly, which may be caused by mild bumping or irregular shapes of the sawn lumber. Figures~\ref{fig:1c}~and~\ref{fig:1d} are the cases of severe distortions, where pixel misalignment is distinct and consistent across the entire piece of lumber. Common defects such as knots or cracks appear in similar colors to the background in the scanned images. They are most frequently found on edges of the lumber specimen. Figure~\ref{fig:examples} also shows that the background color of the raw images is not uniformly black. 

These features of raw images make it a challenging task to align the pixels simply by shifting the columns of pixels based on a given threshold. In the preliminary analysis, we experimented with the method based on threshold to fix the pixel misalignment in the raw images. We first transform the color images to grey scale. For each column of pixels, we then find the position of the first pixel has values greater than the threshold and shift the entire column to offset the difference. 

\begin{figure}[ht]
	\centering
	\begin{minipage}[b]{\linewidth}
		\begin{subfigure}{\textwidth}
			\centering
			\includegraphics[width=0.9\linewidth]{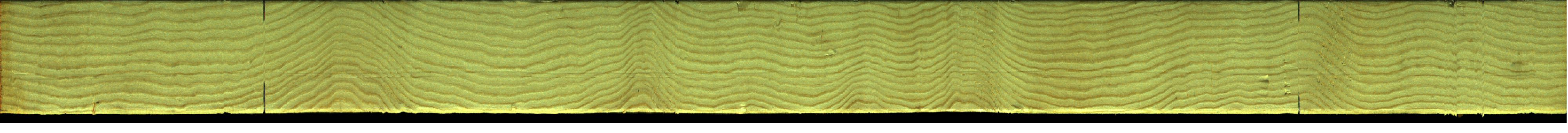}
		\end{subfigure}
		\subcaption{Moderate distortion with no or minor defects}\label{fig:2a}
		
		\begin{subfigure}{\textwidth}
			\centering
			\includegraphics[width=0.9\linewidth]{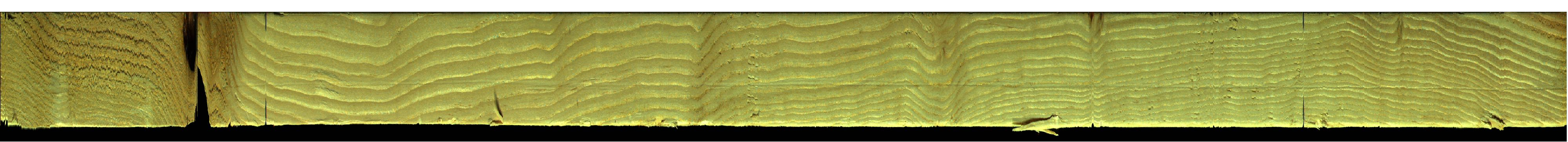}
		\end{subfigure}
		\subcaption{Moderate distortion with evident defects}\label{fig:2b}
		
		\begin{subfigure}{\textwidth}
			\centering
			\includegraphics[width=0.9\linewidth]{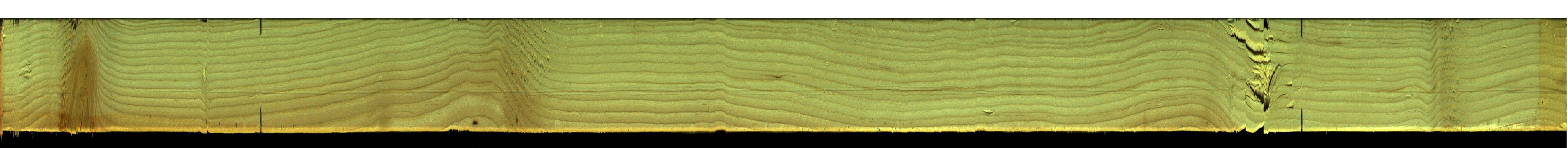}
		\end{subfigure}
		\subcaption{Severe distortion with no or minor defects}\label{fig:2c}
		
		\begin{subfigure}{\textwidth}
			\centering
			\includegraphics[width=0.9\linewidth]{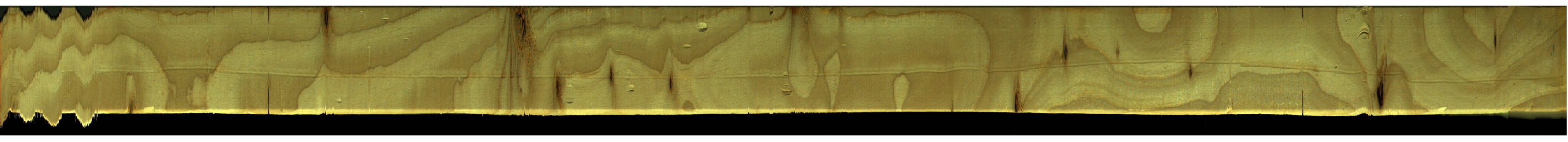}
		\end{subfigure}
		\subcaption{Severe distortion with evident defects}\label{fig:2d}
	\end{minipage}
	\caption{Results of images processed using the method based on threshold.}
	\label{fig:naive_threshold}
\end{figure}

This method works reasonably well for specimens with almost clear surfaces as shown in Figure~\ref{fig:2a}. However, it is not robust to the presence of common lumber defects such knots or cracks. In Figure~\ref{fig:2b}, the pixel misalignment is not correctly fixed, especially around knot areas where the color of the knot is similar to the background color. In Figure~\ref{fig:2c}, the alignment fails around the cracks. Moreover, due to the non-uniform background color, the threshold method may fail and generate images with non-smooth edges as shown in Figure~\ref{fig:2d}.

\begin{figure}[ht]
	\centering
	\begin{minipage}[b]{\linewidth}
		\begin{subfigure}{\textwidth}
			\centering
			\includegraphics[width=0.9\linewidth]{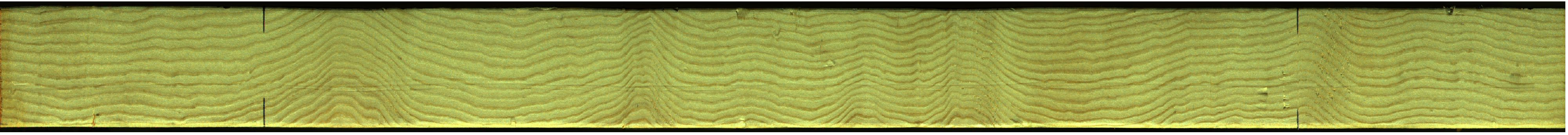}
		\end{subfigure}
		\subcaption{Moderate distortion with no or minor defects}\label{fig:3a}
		
		\begin{subfigure}{\textwidth}
			\centering
			\includegraphics[width=0.9\linewidth]{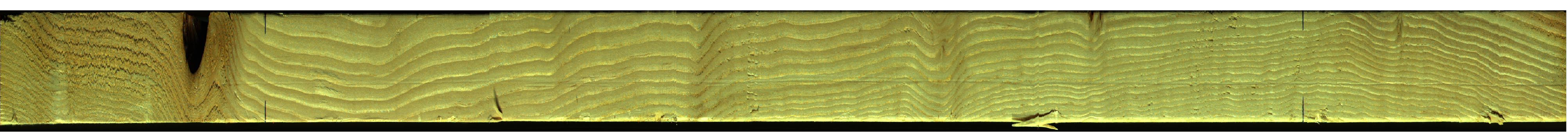}
		\end{subfigure}
		\subcaption{Moderate distortion with evident defects}\label{fig:3b}
		
		\begin{subfigure}{\textwidth}
			\centering
			\includegraphics[width=0.9\linewidth]{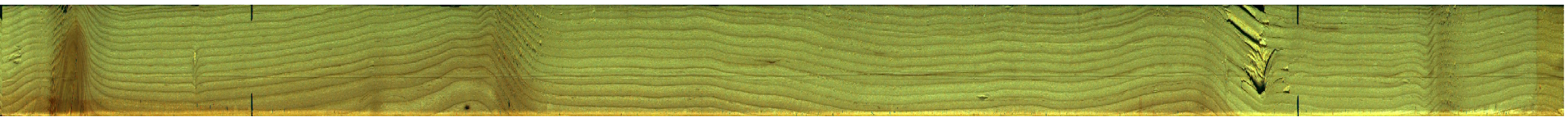}
		\end{subfigure}
		\subcaption{Severe distortion with no or minor defects}\label{fig:3c}
		
		\begin{subfigure}{\textwidth}
			\centering
			\includegraphics[width=0.9\linewidth]{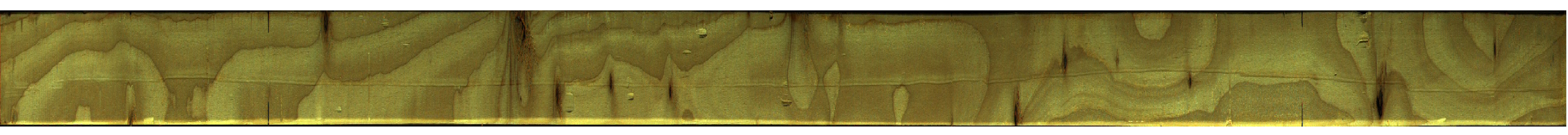}
		\end{subfigure}
		\subcaption{Severe distortion with evident defects}\label{fig:3d}
	\end{minipage}
	\caption{Results of images processed using our proposed method.}
	\label{fig:result_algorithm}
\end{figure}

In contrast, our proposed method is more robust to the presence of common defects as well as non-uniform background color. Figure~\ref{fig:result_algorithm} shows the images processed using our proposed method in the four cases.

{\small
\bibliographystyle{ieee_fullname}
\bibliography{egbib}
}

\end{document}